\relax
\documentclass[letterpaper]{article} 
\usepackage{aaai19}  

\usepackage{times}  
\usepackage{helvet} 
\usepackage{courier}  
\usepackage[hyphens]{url}  
\usepackage{graphicx} 
\usepackage{graphicx}  
\usepackage{import}
\usepackage{tikz}

\title{Learning Norms from Stories: A Prior for Value Aligned Agents}

\newcommand{\GG}{\textit{Goofus \& Gallant}}
\newcommand{\GnG}{\textit{G\&G}}
 \author{Spencer Frazier\thanks{Equal contribution.}\\
{Georgia Institute of Technology}\\
sfrazier7@gatech.edu
\And
Md Sultan Al Nahian\textsuperscript{*}\\
{University of Kentucky}\\
sa.nahian@uky.edu
\And
Mark Riedl\\
{Georgia Institute of Technology}\\
riedl@cc.gatech.edu
\And
Brent Harrison\\
{University of Kentucky}\\
harrison@cs.uky.edu}
\usepackage{array}
\usepackage{url}

\begin{document}

\maketitle
\begin{abstract}
Value alignment is a property of an intelligent agent indicating that it can only pursue goals and activities that are beneficial to humans.
Traditional approaches to value alignment use imitation learning or preference learning to infer the values of humans by observing their behavior.
We introduce a complementary technique in which a value-aligned prior is learned from naturally occurring stories which encode societal norms. Training data is sourced from the children's educational comic strip, \GG{}.
In this work, we train multiple machine learning models to classify natural language descriptions of situations found in the comic strip as normative or non-normative by identifying if they align with the main characters' behavior.
We also report the models' performance when transferring to two unrelated tasks with little to no additional training on the new task.


\end{abstract}

\section{Introduction}

{\em Value alignment} is a property of an intelligent agent indicating that it can only pursue goals and activities which are beneficial to humans~\cite{soares2014aligning,russell2015research,arnold2017value}.
Russell~\shortcite{russell-new-book}, Moor~\cite{moor2006nature}, and others have argued that value alignment is one of the most important tasks facing AI researchers today.
Ideally, a value-aligned system should make decisions that
align with human decisions in similar situations 
and, in theory, make decisions which are unlikely to be harmful~\cite{bostrom}.

Value  alignment, unfortunately, is  not  trivial  to  achieve.  
As  articulated  by Soares~\shortcite{soares2015value}, it is very hard to directly specify values because  there  are  infinitely  many undesirable outcomes in an open world. 
Thus, a sufficiently intelligent artificial agent can unintentionally violate the intent of the tenants of a behavioral rule set without explicitly violating any particular rule.
Recently, approaches to value alignment have largely relied on learning from observations or other forms of imitation learning ~\cite{stadie2017third,Wulfmeier2019EfficientSF,ho2016generative}.
Values can be cast as {\em preferences} over action sequences; 
preference learning can be formulated as reward learning or imitation learning ~\cite{russell-new-book}. 
The difficulties with value alignment via imitation learning are threefold:
(1)~learning knowledge from demonstrations that generalizes beyond the context of the observation is difficult;
(2)~it can be time consuming to provide sufficient demonstrations and, if the agent is learning online, it can be performing harmful actions until learning is complete;
and lastly
(3)~it can be difficult for humans to provide high quality demonstrations that exemplify certain values, especially those related to negation or {\em not} doing something.

In situations where imitation learning is difficult to achieve---such as those above---we propose that a strong prior belief over the quality of certain actions or events can complement imitation learning-based approaches.
A strong prior for value-aligned actions may replace the need for imitation learning or, more likely, make it easier for an imitation learner to align itself with values.
%
From where can we acquire this strong prior?
One solution is to learn this prior through stories ~\cite{harrison2016learning}. 
Stories contain examples of normative and non-normative behavior~\cite{riedl-hcml2016}. We define normativity as behavior which conforms to expected societal norms and contracts whereas non-normativity aligns to values which deviate from these expected norms. Non-normativity does not connotate behavior devoid of value.
Some examples of stories designed to explicitly teach normative behavior are children's literature, allegorical tales, and Aesop's fables. 
Stories for entertainment can also contain examples of normative and non-normative behavior.
Protagonists often exemplify the virtues that a particular culture or society idealize, while antagonists regularly violate one or more social norms.

\begin{figure}
\centering{
\includegraphics[width=0.95\columnwidth]{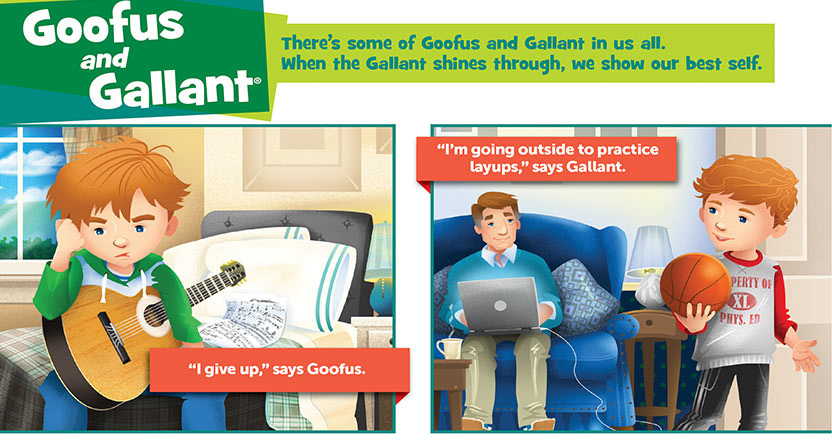}} 
\caption{A modern example of \textit{Goofus \& Gallant} }
\label{fig:gg}
\end{figure}

We explore how a strong prior can be best learned from naturally occurring story corpora.
First, one must be able to reason about the context of individual sentences. 
We turn to language modeling techniques that can extract contextual semantics from sentences. 
Second, there is presently a lack of readily available, labeled datasets with normative behavior descriptions to train on.
Despite the general prevalence of stories in society, stories very rarely explicitly outline values or social norms present in them.
A reasonable starting point is to focus on children's stories that are meant to teach through examples of normative behavior.
Specifically, we have identified a children's cartoon called \GG{} (\GnG).
The cartoon features two characters, Goofus and Gallant, in common everyday scenarios, such that Gallant always acts ``properly'' and Goofus always performs some action that would be considered ``improper'' at that moment 
(see Figure~\ref{fig:gg}).
The \GG{} dataset can thus be thought of as a labeled dataset of normative behavior descriptions. 

In this paper, we describe how we learn a value-aligned prior from the naturally occurring \GG{} corpus.
We show that we can learn to classify sentences from \GG{} as normative or non-normative with a high degree of accuracy.
However, that tells us little about whether such a model can act as a prior for other tasks for which there is no labeled data about normative behavior.
We further show that our model trained on \GnG{} performs adequately at zero-shot transfer when classifying behavior in corpora for which there are no ground-truth normative labels. Since zero-shot transfer is done without additional training on the new task, we have evidence that the dataset and model can act as a value-aligned prior over behavior descriptions.
With some small amount of labeled data in the new task, the prior becomes nearly as strong as when the model is used to classify \GnG{} sentences.

The \GnG{} dataset implies that we are only modeling Western (specifically American) values.
However, values can be aligned to other cultures and societies should analogous datasets be identified and used.
We discuss the ethical implications of our work at the end of this paper.


\section{Related Work}

Humans have expectations that---just like other humans---agents will conform to personal values and to social norms~\cite{bicchieri2005grammar}, even when not explicitly communicated. 
This is the value alignment problem~\cite{soares2014aligning,russell2015research,taylor2016alignment,arnold2017value,abel2016reinforcement}. 
Some assert that agents should be imbued with the capability for moral decision making~\cite{dehghani2008integrated,sun2013moral}, but morals are more difficult to define than values or norms. 
Values themselves are not so simple to define~\cite{soares2015value} and grappling with the philosophical debate over values is out of the scope of this paper. 

Some approaches to value alignment include learning from expert demonstrations ~\cite{schaal1997learning,ho2016showing}, preference learning ~\cite{akrour2012april,christiano2017deep}, imitation ~\cite{ho2016generative} and inverse reinforcement learning~\cite{ng2000algorithms}. 
Cooperative inverse reinforcement-learning ~\cite{hadfield2016cooperative}, for example, works to derive the reward function exhibited by a human for some task. 
These methods are costly in terms of the amount of human input required to train the model. 
These approaches assume that values are latent within people but can be teased out in the form of a reward from which an agent can learn.
As with any problem with a sparse or expensive to acquire signal, 
there is a need for a strong prior to assure transferability ~\cite{zoph2016transfer}.

%

Learning from Stories  ~\cite{riedl2016using,harrison2016learning} is similar to learning from demonstration, except the demonstrations are replaced by natural language stories; a reinforcement learning agent extracts reward signal from the stories to perform more human-like action sequences.
It was shown that agents could learn to avoid non-normative behavior whenever possible. 
Learning from Stories (LfS) is the first attempt at value iteration in reinforcement learning using story content. 
However, the stories used were crowdsourced instead of using a naturally occurring corpus and thus still expensive.
Our work differs by focusing on value alignment as a prior instead of directly learning a value-aligned policy. 
Our work complements LfS and other approaches involving learning from demonstration or imitation learning by providing a means of {\em a priori} biasing the agent toward certain actions.
%

The most similar work is that by Ziegler et al.~\shortcite{ziegler2019} in which the transformer-based language model, GPT-2, is fine-tuned to learn preferences for generating sentences. 
While sentiment is not the same as values, it shows that language models can be trained from human preference data.



\section{Datasets}

We describe the \GG{} (\GnG) training corpus, a source of textual descriptions of everyday life situations and ground-truth labels of normative and non-normative behavior.
In order to show transfer of models trained on \GnG{} transfer to other tasks, we collect two other datasets of situation descriptions, which are labeled via crowdsourcing. 

\subsection{\GG}

It is difficult to curate a corpus of naturally occurring stories for the purposes of learning social norms because authors often assume that the reader has this knowledge. 
Children's stories, however, can prove useful as they are often used as tools to impart knowledge of social conventions, values, and other cultural knowledge to our children. 
In order for a story to be suitable for use in training our machine learning models, however, there must be a way to easily extract labels of normative and non-normative behavior. 
We introduce the \GG{} (\GnG) corpus, composed of excerpts taken from the popular children's comic strip of the same name.
\GG{} (Figure~\ref{fig:gg}) is a children's comic strip that has appeared in the U.S. children's magazine, {\em Highlights}, since 1940. 
It features two main characters, Goofus and Gallant, who are depicted in common everyday scenarios that young children might find themselves in. 
These comics are meant to illustrate the proper way to navigate a situation and the improper way to navigate the situation based on which character is performing the action. 
Gallant is meant to act ``properly'' or in a socially acceptable way, whereas Goofus is meant to navigate the situation ``improperly'' or in a way that violates social conventions or norms. 
For our purposes, \GnG{} is an ideal story corpus; normative behavior is tightly coupled with behaviors associated with the character Gallant. 
The presence of Goofus ensures that we have negative examples that are identified as such. 

\GnG{} comics have been being released monthly since 1940, meaning that the social conventions portrayed in these comics have evolved greatly since their inception. 
To better ensure that our machine learning models learn relevant social norms, we have curated a corpus of \GnG{} comics that consist only of recent comics from 1995 to 2017. 
Since we only use text to train our model, we extract only the text from each comic panel. We then remove explicit references to Goofus and Gallant by replacing their names with pronouns like ``he'', ``she'', or ``they''. 
%
Goofus always portrays an antagonist character doing only socially unacceptable actions. Gallant portrays a protagonist character doing socially acceptable actions. We treat the opposing panes as labels. 
All actions done by Goofus are labeled negative and all the actions done by Gallant labeled as positive.
This provides us with 1,387 sentences.
For all of the experiments in this paper, we use a training set consisting of 50\% of the corpus and a test set of the remaining 50\% of the corpus. 

\subsection{{\em Plotto} Dataset}

\textit{Plotto} ~\cite{plotto} is a book written to help provide inspiration and guidance to potential writers by providing a large library of thousands of predetermined narrative events, called {\em plot points}, commonly found in fiction. 
By expounding on one of the primary theories of storytelling---``\textit{Purpose}, opposed by \textit{obstacle}, yields \textit{conflict}''---thousands of branching situations and scenarios are presented. 
Within each plot point there are one or more character slots with one character always being the primary actor/actress. 
This text provides us with a large number of potential story events to test our models' performance. 
The corpus was extracted from the book with the aide of open-source software described in~\cite{eger2018dairector}. 

In \textit{Plotto} there are 1,462 plot points provided. 
This book was originally published in 1928 and contains several plot events which are overtly racist or misogynistic.
For our experiments, we removed these plot events, which reduced the total number of plot points available from 1,462 to 900. 

To test transfer on this dataset, we require normative/non-normative labels for each plot event. 
We crowdsourced labels via TurkPrime~\cite{litman2017turkprime}, a service which manages Amazon Mechanical Turk tasks with US-based workers.
We designed a survey in which participants are asked to label each phrase extracted from  \textit{Plotto} plot points as normative or non-normative. 
Specifically, we prompt the individuals labeling to consider whether the behavior would be surprising or unsurprising given the context. 
$N=5$ classifications were obtained for each plot point. 
Plot points receiving more than one dissenting classification were discarded, and the remaining ones were given a label based tagged consensus. 
After this process, the corpus contained 555 phrases subsequently used in our transfer experiments.

\subsection{Science Fiction Summaries Dataset}

To further test the transfer capabilities of our trained machine learning models, we used a second, open-source dataset composed of plot summaries taken from fan wikis for popular science fiction shows such as \textit{Babylon 5}, \textit{Dr. Who}, and \textit{Star Trek}, and movies such as \textit{Star Wars}~\cite{Ammanabrolu2019StoryRE}.
%
In this corpus, we make the assumption that each sentence encodes at least one plot event in the overall story. 
First, we manually extracted sentences containing character-driven events.
During this process, we identified that some sentences actually encode multiple events and contain both normative and non-normative behaviors. 
In these cases, we manually divided the sentence into multiple separate events. 
After this manual extraction, this corpus contained 800 story events. 
As with the \GnG{} dataset, We replace common character names such as Anakin, Skywalker, or Darth Sidious with pronouns.

To label plot events in this corpus, we followed a procedure similar to that used to tag the \textit{Plotto} dataset. 
Participants were asked to consider normativity within the context of the science fiction universe that the event takes place in. 
This is to avoid situations where actions are labeled as being non-normative due to discrepancies between the real world and the science fiction world. 
As with the {\em Plotto} dataset, we obtain $N=5$ classifications for each summary sentence and discard any sentences for which there was at least one dissenting vote. 
After this process, our science fiction corpus contained 445 annotated sentences with consensus. 
A summary of each dataset used in our experiments can be found in Table~\ref{tab:tabledata}.

\begin{table}[tb]
\centering
\footnotesize
\caption{Dataset summaries.} 
\label{tab:tabledata}

\begin{tabular}{l|c|c|c}
\multicolumn{1}{c}{\textbf{Dataset}} &
\multicolumn{1}{c}{\textbf{Original N}} &
\multicolumn{1}{c}{\textbf{Hand-Selected N}} &
\multicolumn{1}{c}{\textbf{Consensus N}} \\
\hline
G\&G
&$1387$&$1387$&N/A\\
\textit{Plotto}
&$1462$&$900$&$555$\\
Sci-Fi
&$4592$&$800$&$445$\\
\hline
\end{tabular}
\end{table}

\begin{figure}
\centering{
\includegraphics[width=1.0\columnwidth]{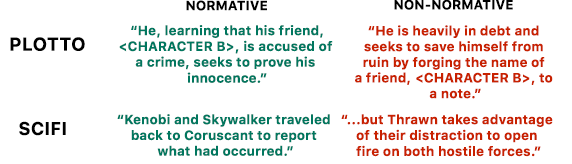}} 
\caption{Examples of test dataset text. }
\label{fig:nvnn}
\end{figure}


\section{Methods}

We seek to show that a model trained on a dataset of normative behavioral natural language examples can 
(a)~identify socially normative behavior and 
(b)~transfer that knowledge to previously unseen examples of behavior. 
In doing so, we are testing our hypothesis that stories contain a great deal of knowledge about sociocultural norms that reflect the society and culture from which the stories were written that can be generalized to different situations. 
We conduct two experiments. 
The first experiment seeks to determine the best machine learning technique for producing a classification model for descriptions normative and non-normative events.
This is done by training several ML models on the \GnG{} training corpus and then measuring classification accuracy on the \GnG{} testing set.
In the second experiment, we explore how the trained model from the first experiment can  transfer to other, unrelated story domains with various amounts of fine-tuning. 
For this experiment, we use the models trained on the \GnG{} corpus to classify events in the \textit{Plotto} dataset and the science fiction summary datasets. 

\subsection{Models}

Using the text of the \GnG{} corpus, we have trained binary classifiers which can classify events in story as normative or non-normative. 
The classifiers take a single sentence as input and the output is whether the sentence contains normative behavior or a non-normative behavior.
We used four different machine learning techniques to build the classifiers: 
(1)~Bidirectional LSTM,
(2)~Deep Pyramid CNN,
(3)~BERT and 
(4)~XLNet.

The Bidirectional LSTM (BiLSTM)~\cite{huang2015bidirectional} works as follows.
An input sentence is encoded using bidirectional multilayer LSTM cell having 2 layers with a size of 512. 
Pretrained GloVe~\cite{glove2014} word embeddings are used to embed the input sentence before passing it through the LSTM layer. 
The hidden state of the LSTM layer is passed through a fully connected (FC) layer followed by a classification layer to make the label prediction. 
The dimension of the FC layer is 4H x 512 and classification layer is 512 x K, where H is the hidden state size of LSTM cell which is 512 and K is the number of classes. 

Using sentiment as a classification signal is a common strategy for performing binary classification on text corpora. 
Deep Pyramid CNNs (DPCNN)~\cite{johnson-zhang-2017-deep} were originally designed for sentiment classification and achieved state-of-the-art sentiment classification results, so we explore how they perform on identifying normative behavior.
A simple network architecture achieves the best accuracy with 15 weight layers.
We re-trained DPCNN on the \GnG{} dataset. 
No pretrained word embeddings were used as the network applies text region embeddings enhanced by unsupervised embeddings ~\cite{cnnregionembed2015}.

BERT~\cite{devlin2018bert} is a transformer that makes use of an attention mechanism to learn contextual relations between words (or sub-words) in a text. 
It achieves strong results on many tasks through its bidirectionality, enabled by token masking.
We utilize BERT's binary classification mode. 
The [CLS] token is omnipresent within the BERT model but only active for classification.
The final hidden state of the [CLS] token is taken as the pooled representation of the input text. 
This is fed to the classification layer which has a dimension of H x K, where K is the number of classes and H is the size of the hidden state. 
Class probabilities are computed via softmax. 

XLNet~\cite{xlnet2019} is a generalized autoregressive pretraining model based on the state-of-the-art autoregressive language model Transformer-XL~\cite{transformer-xl2019}, which removes MASK tokens while incorporating permutation language modeling to capture the bidirectional context. 
We utilize XLNet for classification by following the same procedure used for BERT.

\begin{table*}[tb]
\centering
\footnotesize
\caption{Results for \GG{} classification experiments. }
\label{gg-training}
\begin{tabular}{l|c|c|c|c|c|c}
\multicolumn{1}{c}{\textbf{Model}} &
\multicolumn{1}{c}{\textbf{Test acc}} &
\multicolumn{1}{c}{\textbf{$F_1$-score}} & 
\multicolumn{1}{c}{\textbf{Precision}} & 
\multicolumn{1}{c}{\textbf{Recall}} &
\multicolumn{1}{c}{\textbf{MCC}} \\
\hline\hline
Human (N=20)
&0.818&0.839&0.925&$0.768$&$0.277$\\
\hline
Bi-LSTM
&$0.687$&$0.674$&$0.729$&$0.687$&$0.417$\\
DPCNN
&$0.754$&$0.748$&$0.784$&$0.754$&$0.538$\\
BERT-Base
&$0.614$&$0.501$&$0.731$&$0.381$&$0.267$\\
XLNet-Base
&$0.606$&$0.585$&$0.628$&$0.547$&$0.214$\\
BERT-GG
&\textbf{0.908}&\textbf{0.907}&\textbf{0.931}&\textbf{0.885}&\textbf{0.818}\\
XLNet-GG
&0.846&$0.834$&$0.918$&$0.765$&$0.702$\\
\end{tabular}
\end{table*}

\subsection{Experimental Setup}

The Bi-LSTM and DPCNN are trained on the \GnG{} training set.
We produced several versions of BERT and XLNet models: 
BERT-Base and XLNet-Base receive no training on \GnG, 
while
BERT-GG and XLNet-GG are fine-tuned on the \GnG{} training set.
All models are tested on a held-out testing set.
For experiment 2, the Bi-LSTM-Plotto/scifi and the DPCNN-Plotto/scifi were first trained \GnG{} and then fine-tuned on the \textit{Plotto} and science fiction datasets respectively. 

Metrics used to evaluate the models include: accuracy, precision ($\frac{TP}{TP+FP}$), recall ($\frac{TP}{TP+FN}$),  $F_1$-score and classification quality as determined by the Matthews correlation coefficient (MCC). 



\begin{table*}[tb]
\centering
\footnotesize
\caption{Results for \textit{Plotto} transfer experiments. 
The BERT-Plotto and XLNet-Plotto models were first trained on \GnG{} and then additionally trained on the Plotto corpus. }
\label{plotto-fine-tuning}

\begin{tabular}{l|c|c|c|c|c}
\multicolumn{1}{c}{\textbf{Model}} &
\multicolumn{1}{c}{\textbf{Test acc}} &
\multicolumn{1}{c}{\textbf{$F_1$-score}} & \multicolumn{1}{c}{\textbf{Precision}} & \multicolumn{1}{c}{\textbf{Recall}} &
\multicolumn{1}{c}{\textbf{MCC}} \\
\hline\hline
Bi-LSTM
&$0.636$&\textbf{0.67}&\textbf{0.735}&$0.636$&$0.146$\\
DPCNN
&$0.525$&$0.555$&$0.645$&$0.525$&$0.058$\\
BERT-Base
&$0.529$&$0.402$&$0.297$&$0.619$&$0.103$\\
XLNet-Base
&$0.46$&$0.436$&$0.297$&$0.817$&$0.148$\\
BERT-GG
&\textbf{0.741}&$0.514$&$0.494$&$0.535$&\textbf{0.338}\\
XLNet-GG
&$0.543$&$0.506$&$0.349$&\textbf{0.915}&$0.307$\\
\hline
Bi-LSTM-Plotto
&$0.737$&\textbf{0.655}&$0.661$&$0.737$&$0.064$\\
DPCNN-Plotto
&0.748&$0.644$&\textbf{0.812}&\textbf{0.748}&$0.103$\\
BERT-Plotto
&\textbf{0.838}&$0.634$&$0.75$&$0.549$&$0.544$\\
XLNet-Plotto
&\textbf{0.838}&$0.651$&$0.724$&$0.592$&\textbf{0.552}
\end{tabular}
\end{table*}

\begin{table*}[tb]
\centering
\footnotesize
\caption{Results for science fiction summary transfer experiments. 
The BERT-scifi and XLNet-scifi models were first trained on \GnG{} and then additionally trained on the Sci-Fi corpus.}
\label{scifi-fine-tuning}
\begin{tabular}{l|c|c|c|c|c|c}
\multicolumn{1}{c}{\textbf{Model}} &
\multicolumn{1}{c}{\textbf{Test acc}} &
\multicolumn{1}{c}{\textbf{$F_1$-score}} & \multicolumn{1}{c}{\textbf{Precision}} & \multicolumn{1}{c}{\textbf{Recall}} &
\multicolumn{1}{c}{\textbf{MCC}} \\
\hline\hline

Bi-LSTM
&$0.511$&$0.519$&$0.54$&$0.511$&$0.015$\\
DPCNN
&$0.521$&$0.528$&$0.558$&$0.52$&$0.052$\\
BERT-Base
&$0.43$&$0.38$&$0.6$&$0.279$&$-0.037$\\
XLNet-Base
&$0.538$&$0.599$&$0.658$&$0.55$&$0.066$\\
BERT-GG
&$0.65$&$0.655$&$0.86$&$0.529$&$0.381$\\
XLNet-GG
&\textbf{0.731}&\textbf{0.784}&\textbf{0.79}&\textbf{0.779}&\textbf{0.427}\\
\hline
Bi-LSTM-scifi
&$0.641$&$0.632$&$0.629$&$0.641$&$0.204$\\
DPCNN-scifi
&$0.646$&$0.531$&$0.712$&$0.646$&$0.159$\\
BERT-scifi
&\textbf{0.874}&\textbf{0.895}&\textbf{0.94}&$0.85$&\textbf{0.747}\\
XLNet-scifi
&0.839&$0.87$&$0.882$&\textbf{0.857}&$0.658$\\
\end{tabular}
\end{table*}

\subsection{Experiment 1: \GG{} Classification}
In the first study, we seek to understand how well a model can classify previously unseen \GnG{} scenarios when trained explicitly on a \GnG{} training set. 
This gives us a base understanding of how well machine learning models can identify information about social norms from story corpora.

The Bi-LSTM network was trained for 80 epochs and the DPCNN was trained for 20 epochs. Both used Adam optimizer and a learning rate of 0.001.
Fine-tuning for the BERT-GG and XLNet-GG models was done using the following parameters:
maximum sequence length of 128 characters, 1 gradient accumulation step, and the learning rate is 4e-5. Model performance peaked at 6 epochs.


Additionally, we conducted a human participant study to determine human accuracy on the task of classifying \GnG{} events as normative or non-normative. 
The study used the same protocol that was used to label the {\em Plotto} and Sci-Fi corpora.
$N=20$ participants tagged sentences from \GG{} and we compared their tags to the ground truth from the original cartoons.

Experiment results for case study 1 are given in Table \ref{gg-training}.
First, it shows that humans have strong agreement with the \GnG{} ground truth labels.
Among the non-transformer models, DPCNN better classifies normative and non-normative behavior from the \GnG{} dataset. 
This is likely because the CNN can identify the global sentence structure better than a simple bi-directional LSTM cell.
While the BERT-Base and XLNet-Base models struggle to classify events from the \GnG{} corpus (achieving accuracies of \%61.4 and \%60.6 respectively), fine-tuning drastically improves each model's performance. 
BERT-GG obtains the best results in each of our metrics, obtaining a $21.33\%$ accuracy improvement over the DPCNN. 

The fine-tuned transformer models share many traits with CNNs in their ability to identify the global context of a sequence of text. 
Additionally, the contextualized word embeddings used in transformer-based models allow for words to have different vector representations based on context, whereas the embeddings used in the non-transformer approaches will often have the same word embedding regardless of context. 
This property is particularly important for our task as many actions in stories can have different meanings based on the situation. 

\subsection{Experiment 2: Transfer}

In this experiment, we investigate how well machine learning models trained to identify normative and non-normative behavior in the \GnG{} corpus can transfer to other story domains. 
Specifically, we explore how well these models can classify events from the \textit{Plotto} and science fiction summary corpora. 
We evaluate how well these models perform on fine-tuned and zero-shot transfer learning.
Fine-tuned transfer learning means using a model trained for one task on a different, but related, task utilizing some additional training for fine-tuning. 
Zero-shot transfer, however, involves using the previously trained model on the new task with no additional training. 
%
Zero-shot transfer is important for use cases where a value-aligned classification model is acquired by training on an unrelated dataset (such as \GnG) and applied to a different task because it is likely that ground truth data on values will not be available to use for additional training.
If some labeled data associated with the new task can be acquired, however, then a fine-tuning transfer protocol can be used.
%

\subsubsection{\GnG{} to Plotto Transfer}

Table \ref{plotto-fine-tuning} shows the results of transfer learning for the \textit{Plotto} dataset. 
Zero-shot transfer results are achieved by testing Bi-LSTM, DPCNN, BERT-GG and XLNet-GG on the {\em Plotto} dataset; these models were trained on \GnG{} but have never seen Plotto plot events. 
BERT-GG outperforms all the other models in the zero-shot transfer in terms of accuracy and MCC. 
These results demonstrate that the knowledge of normative and non-normative behavior gathered from the \GnG{} stories alone facilitates a strong prior over normative/non-normative behavior without overfitting to \GnG{} scenarios and language. 


To further investigate the transferability of the models, we fine tuned all the \GnG{} models (Bi-LSTM, DPCNN, BERT-GG and XLNet-GG) on {\em Plotto} stories. 
When fine-tuning each model, we use the same parameter settings used in experiment 1 except for the number of training epochs. 
We fine-tuned the Bi-LSTM-Plotto for 20 epochs, DPCNN-Plotto for 4 epochs, BERT-Plotto and XLNet-Plotto for 3 epochs. Epoch count for transformers is low due to their propensity to overfit and lose the advantage of their pretrained weights.

%
Results from the experiment show that fine-tuning these models on the Plotto dataset significantly increases model performance.
Even though all model performance increases, the transformer models still drastically outperform both non-transformer methods. 

\subsubsection{\GnG{} to Sci-Fi Transfer}

Events in \GnG{} stories are from our daily life whereas Sci-Fi plots are fictional, consisting of strange objects and events. We use the science fiction plot summary dataset to show the capability these models have for transfer learning in another narrative context.
The results for this second experiment are shown in Table ~\ref{scifi-fine-tuning}. 
As before, we find that transformer-based models perform well on zero-shot transfer, though in this case they perform worse than they did with the {\em Plotto} task.
As with the {\em Plotto} task, we also fine-tuned our models on the sci-fi training data using the same training protocol. 
We see a dramatic increase in performance when given access to even a small amount of task-specific normative labels for fine tuning. 



\section{Discussion}

Our experimental results demonstrate that transformer-based models trained on the naturally occurring \GG{} story corpus are highly accurate in classifying previously unseen descriptions of normative behavior taken from that comic strip.
However, a more notable observation is that the best models, the transformer models, can achieve high accuracy when classifying event descriptions from unrelated corpora.
This is significant in that it means the model can transfer to other tasks without requiring any normative/non-normative labels of situations from the new tasks. 
When a small number of labels from the transfer tasks are available, the classification accuracy increases to nearly the same level as when the model is used to classify situations from the \GG{} corpus.

A question that often arises in value alignment research is ``whose values do these models reflect?''.
Our models are trained to classify behavior according to Western (specifically American) cultural norms inherent in these comics. 
Should labeled datasets exhibiting other value systems be identified, our models can be re-trained to reflect those norms instead. 

One limitation of this work is that swapping positive and negative labels would allow an unscrupulous actor to create an anti-value-aligned model.
This model could in turn be used to bias other models to produce non-normative behavior. 
For example, a language generation model such as GPT-2 could be biased in a way that it produces trolling behavior using a technique similar to that in Ziegler et al.~\shortcite{ziegler2019}. 
Likewise, a reinforcement learning agent or robot could be biased toward a non-normative, and thus potentially harmful, action policy.
However, the main use of our work is to complement a more traditional learning by demonstration technique.
A reinforcement learning system biased by an anti-value-aligned prior may be remediated with more demonstrations of normative behavior before converging on a final, value-aligned policy.

Events often have context---the appropriateness of a situation may be conditional on the events that have preceded it.
This is especially true for reinforcement learning agents that learn a sequential task instead of an episodic task.
Another limitation of our models is that they do not currently factor in context that is not present in the sentence being classified.


\section{Conclusions}

Through the use of machine learning, the information contained in stories can be used to learn a strong and robust prior for value alignment. 
This is because characters within stories often embody normative and non-normative behavior. 
By extracting the actions of these characters, story text can be used to train machine learning models that can classify descriptions of normative and non-normative behavior.
In this paper, we introduce the \GG{} corpus, a naturally occurring story corpus with ground truth labels about socially normative and non-normative behaviors.
We show how various machine learning models can be trained on this corpus to produce accurate classifications of behavior and highlight the excellent performance that  transformer-based language models achieve on this task.
We further show that these models can transfer to unrelated event description tasks for which there are no ground truth labels.
Consequently, these models can form a strong prior that complement more traditional value alignment techniques such as learning by demonstration, preference learning, or other forms of imitation learning.


\section*{Acknowledgements}
Highlights for Children, Inc. gave permission for us to use selected \GG{} features to create the dataset described in this paper.


\bibliography{aies_paper}{}

\begin{thebibliography}{}

\bibitem[\protect\citeauthoryear{Abel, MacGlashan, and
  Littman}{2016}]{abel2016reinforcement}
Abel, D.; MacGlashan, J.; and Littman, M.~L.
\newblock 2016.
\newblock Reinforcement learning as a framework for ethical decision making.
\newblock In {\em AAAI Workshop: AI, Ethics, and Society}.

\bibitem[\protect\citeauthoryear{Akrour, Schoenauer, and
  Sebag}{2012}]{akrour2012april}
Akrour, R.; Schoenauer, M.; and Sebag, M.
\newblock 2012.
\newblock April: Active preference learning-based reinforcement learning.
\newblock In {\em Joint European Conference on Machine Learning and Knowledge
  Discovery in Databases},  116--131.
\newblock Springer.

\bibitem[\protect\citeauthoryear{Ammanabrolu \bgroup et al\mbox.\egroup
  }{2019}]{Ammanabrolu2019StoryRE}
Ammanabrolu, P.; Tien, E.; Cheung, W.; Luo, Z.; Ma, W. L.~W.; Martin, L.~J.;
  and Riedl, M.~O.
\newblock 2019.
\newblock Story realization: Expanding plot events into sentences.
\newblock {\em ArXiv} abs/1909.03480.

\bibitem[\protect\citeauthoryear{Arnold, Kasenberg, and
  Scheutz}{2017}]{arnold2017value}
Arnold, T.; Kasenberg, D.; and Scheutz, M.
\newblock 2017.
\newblock Value alignment or misalignment - what will keep systems accountable?
\newblock In {\em AAAI Workshop: AI, Ethics, and Society}.

\bibitem[\protect\citeauthoryear{Bicchieri}{2005}]{bicchieri2005grammar}
Bicchieri, C.
\newblock 2005.
\newblock {\em The grammar of society: The nature and dynamics of social
  norms}.
\newblock Cambridge University Press.

\bibitem[\protect\citeauthoryear{Bostrom}{2014}]{bostrom}
Bostrom, N.
\newblock 2014.
\newblock Superintelligence: Paths, dangers, strategies.

\bibitem[\protect\citeauthoryear{Christiano \bgroup et al\mbox.\egroup
  }{2017}]{christiano2017deep}
Christiano, P.~F.; Leike, J.; Brown, T.; Martic, M.; Legg, S.; and Amodei, D.
\newblock 2017.
\newblock Deep reinforcement learning from human preferences.
\newblock In {\em Advances in Neural Information Processing Systems},
  4299--4307.

\bibitem[\protect\citeauthoryear{Cook}{1920}]{plotto}
Cook, W.~W.
\newblock 1920.
\newblock {\em Plotto: The Master Book of All Plots}.
\newblock Tin House Books.

\bibitem[\protect\citeauthoryear{Dai \bgroup et al\mbox.\egroup
  }{2019}]{transformer-xl2019}
Dai, Z.; Yang, Z.; Yang, Y.; Carbonell, J.~G.; Le, Q.~V.; and Salakhutdinov, R.
\newblock 2019.
\newblock {Transformer-XL}: Attentive language models beyond a fixed-length
  context.
\newblock {\em CoRR} abs/1901.02860.

\bibitem[\protect\citeauthoryear{Dehghani \bgroup et al\mbox.\egroup
  }{2008}]{dehghani2008integrated}
Dehghani, M.; Tomai, E.; Forbus, K.~D.; and Klenk, M.
\newblock 2008.
\newblock An integrated reasoning approach to moral decision-making.
\newblock In {\em AAAI},  1280--1286.

\bibitem[\protect\citeauthoryear{Devlin \bgroup et al\mbox.\egroup
  }{2018}]{devlin2018bert}
Devlin, J.; Chang, M.-W.; Lee, K.; and Toutanova, K.
\newblock 2018.
\newblock Bert: Pre-training of deep bidirectional transformers for language
  understanding.
\newblock {\em arXiv preprint arXiv:1810.04805}.

\bibitem[\protect\citeauthoryear{Eger and Mathewson}{2018}]{eger2018dairector}
Eger, M., and Mathewson, K.~W.
\newblock 2018.
\newblock {dAIrector}: Automatic story beat generation through knowledge
  synthesis.
\newblock {\em CoRR} abs/1811.03423.

\bibitem[\protect\citeauthoryear{Hadfield-Menell \bgroup et al\mbox.\egroup
  }{2016}]{hadfield2016cooperative}
Hadfield-Menell, D.; Russell, S.~J.; Abbeel, P.; and Dragan, A.
\newblock 2016.
\newblock Cooperative inverse reinforcement learning.
\newblock In {\em Advances in neural information processing systems}.

\bibitem[\protect\citeauthoryear{Harrison and
  Riedl}{2016}]{harrison2016learning}
Harrison, B., and Riedl, M.~O.
\newblock 2016.
\newblock Learning from stories: using crowdsourced narratives to train virtual
  agents.
\newblock In {\em Twelfth Artificial Intelligence and Interactive Digital
  Entertainment Conference}.

\bibitem[\protect\citeauthoryear{Ho and Ermon}{2016}]{ho2016generative}
Ho, J., and Ermon, S.
\newblock 2016.
\newblock Generative adversarial imitation learning.
\newblock In {\em Advances in neural information processing systems},
  4565--4573.

\bibitem[\protect\citeauthoryear{Ho \bgroup et al\mbox.\egroup
  }{2016}]{ho2016showing}
Ho, M.~K.; Littman, M.; MacGlashan, J.; Cushman, F.; and Austerweil, J.~L.
\newblock 2016.
\newblock Showing versus doing: Teaching by demonstration.
\newblock In {\em Advances in neural information processing systems},
  3027--3035.

\bibitem[\protect\citeauthoryear{Huang, Xu, and
  Yu}{2015}]{huang2015bidirectional}
Huang, Z.; Xu, W.; and Yu, K.
\newblock 2015.
\newblock Bidirectional lstm-crf models for sequence tagging.
\newblock {\em arXiv preprint arXiv:1508.01991}.

\bibitem[\protect\citeauthoryear{Johnson and Zhang}{2015}]{cnnregionembed2015}
Johnson, R., and Zhang, T.
\newblock 2015.
\newblock Semi-supervised convolutional neural networks for text categorization
  via region embedding.
\newblock In Cortes, C.; Lawrence, N.~D.; Lee, D.~D.; Sugiyama, M.; and
  Garnett, R., eds., {\em Advances in Neural Information Processing Systems
  28}. Curran Associates, Inc.
\newblock  919--927.

\bibitem[\protect\citeauthoryear{Johnson and
  Zhang}{2017}]{johnson-zhang-2017-deep}
Johnson, R., and Zhang, T.
\newblock 2017.
\newblock Deep pyramid convolutional neural networks for text categorization.
\newblock In {\em Proceedings of the 55th Annual Meeting of the Association for
  Computational Linguistics},  562--570.
\newblock Vancouver, Canada: Association for Computational Linguistics.

\bibitem[\protect\citeauthoryear{Litman, Robinson, and
  Abberbock}{2017}]{litman2017turkprime}
Litman, L.; Robinson, J.; and Abberbock, T.
\newblock 2017.
\newblock Turkprime. com: A versatile crowdsourcing data acquisition platform
  for the behavioral sciences.
\newblock {\em Behavior research methods} 49(2):433--442.

\bibitem[\protect\citeauthoryear{Moor}{2006}]{moor2006nature}
Moor, J.~H.
\newblock 2006.
\newblock The nature, importance, and difficulty of machine ethics.
\newblock {\em IEEE intelligent systems} 21(4):18--21.

\bibitem[\protect\citeauthoryear{Ng, Russell, and
  others}{2000}]{ng2000algorithms}
Ng, A.~Y.; Russell, S.~J.; et~al.
\newblock 2000.
\newblock Algorithms for inverse reinforcement learning.
\newblock In {\em Icml}, volume~1, ~2.

\bibitem[\protect\citeauthoryear{Pennington, Socher, and
  Manning}{2014}]{glove2014}
Pennington, J.; Socher, R.; and Manning, C.~D.
\newblock 2014.
\newblock Glove: Global vectors for word representation.
\newblock In {\em Empirical Methods in Natural Language Processing (EMNLP)},
  1532--1543.

\bibitem[\protect\citeauthoryear{Riedl and Harrison}{2016}]{riedl2016using}
Riedl, M.~O., and Harrison, B.
\newblock 2016.
\newblock Using stories to teach human values to artificial agents.
\newblock In {\em Workshops at the Thirtieth AAAI Conference on Artificial
  Intelligence}.

\bibitem[\protect\citeauthoryear{Riedl}{2016}]{riedl-hcml2016}
Riedl, M.~O.
\newblock 2016.
\newblock Computational narrative intelligence: {A} human-centered goal for
  artificial intelligence.
\newblock {\em CoRR} abs/1602.06484.

\bibitem[\protect\citeauthoryear{Russell, Dewey, and
  Tegmark}{2015}]{russell2015research}
Russell, S.; Dewey, D.; and Tegmark, M.
\newblock 2015.
\newblock Research priorities for robust and beneficial artificial
  intelligence.
\newblock {\em Ai Magazine} 36(4):105--114.

\bibitem[\protect\citeauthoryear{Russell}{2019}]{russell-new-book}
Russell, S.~J.
\newblock 2019.
\newblock {\em Human Compatible: Artificial Intelligence and the Problem of
  Con}.
\newblock Viking (October 8, 2019).

\bibitem[\protect\citeauthoryear{Schaal}{1997}]{schaal1997learning}
Schaal, S.
\newblock 1997.
\newblock Learning from demonstration.
\newblock In {\em Advances in neural information processing systems},
  1040--1046.

\bibitem[\protect\citeauthoryear{Soares and
  Fallenstein}{2014}]{soares2014aligning}
Soares, N., and Fallenstein, B.
\newblock 2014.
\newblock Aligning superintelligence with human interests: A technical research
  agenda.
\newblock {\em Machine Intelligence Research Institute technical report} 8.

\bibitem[\protect\citeauthoryear{Soares}{2015}]{soares2015value}
Soares, N.
\newblock 2015.
\newblock The value learning problem.
\newblock {\em Machine Intelligence Research Institute, Berkley}.

\bibitem[\protect\citeauthoryear{Stadie, Abbeel, and
  Sutskever}{2017}]{stadie2017third}
Stadie, B.~C.; Abbeel, P.; and Sutskever, I.
\newblock 2017.
\newblock Third-person imitation learning.
\newblock {\em arXiv preprint arXiv:1703.01703}.

\bibitem[\protect\citeauthoryear{Sun}{2013}]{sun2013moral}
Sun, R.
\newblock 2013.
\newblock Moral judgment, human motivation, and neural networks.
\newblock {\em Cognitive Computation} 5(4):566--579.

\bibitem[\protect\citeauthoryear{Taylor \bgroup et al\mbox.\egroup
  }{2016}]{taylor2016alignment}
Taylor, J.; Yudkowsky, E.; LaVictoire, P.; and Critch, A.
\newblock 2016.
\newblock Alignment for advanced machine learning systems.
\newblock {\em Machine Intelligence Research Institute}.

\bibitem[\protect\citeauthoryear{Wulfmeier}{2019}]{Wulfmeier2019EfficientSF}
Wulfmeier, M.
\newblock 2019.
\newblock Efficient supervision for robot learning via imitation, simulation,
  and adaptation.
\newblock {\em KI - K{\"u}nstliche Intelligenz}  1--5.

\bibitem[\protect\citeauthoryear{Yang \bgroup et al\mbox.\egroup
  }{2019}]{xlnet2019}
Yang, Z.; Dai, Z.; Yang, Y.; Carbonell, J.~G.; Salakhutdinov, R.; and Le, Q.~V.
\newblock 2019.
\newblock Xlnet: Generalized autoregressive pretraining for language
  understanding.
\newblock {\em CoRR} abs/1906.08237.

\bibitem[\protect\citeauthoryear{Ziegler \bgroup et al\mbox.\egroup
  }{2019}]{ziegler2019}
Ziegler, D.~M.; Stiennon, N.; Wu, J.; Brown, T.~B.; Radford, A.; Amodei, D.;
  Christiano, P.; and Irving, G.
\newblock 2019.
\newblock Fine-tuning language models from human preferences.
\newblock {\em arXiv preprint arXiv:1909.08593}.

\bibitem[\protect\citeauthoryear{Zoph \bgroup et al\mbox.\egroup
  }{2016}]{zoph2016transfer}
Zoph, B.; Yuret, D.; May, J.; and Knight, K.
\newblock 2016.
\newblock Transfer learning for low-resource neural machine translation.
\newblock {\em arXiv preprint arXiv:1604.02201}.

\end{thebibliography}
\bibliographystyle{aaai}

\end{document}